\documentclass[10pt,conference]{IEEEtran}
\IEEEoverridecommandlockouts
\usepackage{cite}
\usepackage{amsmath,amssymb,amsfonts}
\usepackage{algorithmic}
\usepackage{graphicx}
\usepackage{textcomp}
\usepackage{xcolor}
\usepackage{multirow}
\def\BibTeX{{\rm B\kern-.05em{\sc i\kern-.025em b}\kern-.08em
    T\kern-.1667em\lower.7ex\hbox{E}\kern-.125emX}}
\begin{document}

\title{On the Identification of the Energy related Issues from the App Reviews\\
\thanks{This study is sponsored by **** Lab of ***** ***** University, **********}
}

\author{\IEEEauthorblockN{ Noshin Nawal}
\IEEEauthorblockA{\textit{Department of Computer Science} \\
\textit{BRAC University}\\
Dhaka, Bangladesh \\
noshin.nawal@bracu.ac.bd}
}

\maketitle


\begin{abstract}
  The energy inefficiency of the apps can be a major issue for the app users which is discussed on App Stores extensively. Previous research has shown the importance of investigating the energy related app reviews to identify the major causes or categories of energy related user feedback. However, there is no study that efficiently extracts the energy related app reviews automatically. 
  
  In this paper, we empirically study different techniques for automatic extraction of the energy related user feedback. We compare the accuracy, F1-score and run time of numerous machine-learning models with relevant feature combinations and relatively modern Neural Network-based models. In total, 60 machine learning models are compared to 30 models that we build using six neural network architectures and three word embedding models. We develop a visualization tool for this study through which a developer can traverse through this large-scale result set.
  The results show that neural networks outperform the other machine learning techniques and can achieve the highest F1-score of 0.935. To replicate the research results, we have open sourced the interactive visualization tool.
  
  After identifying the best results and extracting the energy related reviews, we further compare various techniques to help the developers automatically investigate the emerging issues that might be responsible for energy inefficiency of the apps. We experiment the previously used \textit{string matching} with results obtained from applying two of the state-of-the-art topic modeling algorithms, OBTM and AOLDA. 
  Finally, we run a qualitative study performed in collaboration with developers and students from different institutions to determine their preferences for identifying necessary topics from previously categorized reviews, which shows OBTM produces the most helpful results.
\end{abstract}


\section{Introduction}
    Energy efficiency in mobile applications is a crucial concern for app-developers, as most of the energy related issues are usually identified from users' feedback after the publication of the app\cite{Oliner13}.
    Energy inefficiency, also referred to as energy consumption or battery consumption in mobile apps, was a main concern in 2013\cite{wilke2013energy} and is still a main topic that appears in user reviews in 2019\cite{noei2019too}.
    
    Energy-related user reviews can contain critical information about the energy inefficiency of the app. Consider an example that is written for app \textit{Trucker Tool}: "GPS running on background uses alot of battery. It kills alot of power of my droid. had to uninstall."
    This review contains useful information about how a specific functionality of the app (Geo-location tracking) led to power related concerns. Investigation into these reviews will allow the developers to identify different aspects and features of the app responsible for the energy related issues\cite{Li14}.
    
    As we will discuss in the paper, finding these insightful feedbacks amongst the hoard of reviews (hundreds or thousands of them) submitted by the users for an app is not an easy task.
    Similar examples suggest that it is imperative for the developers to efficiently extract energy related reviews. 
    However, to the best of our knowledge, there is no study that investigates the automatic extraction of energy consumption reviews. 
    
    In this paper, we empirically study 60 traditional machine learning models with various sets of feature combinations and 30 deep learning models (having six different architectures) with three pretrained word embeddings to identify the best features- model combination for better extraction of energy related app reviews from Google Play store, in supervised text classification approach. The models are executed on the reviews scraped for 400 apps from 12 different categories. 
    As the number of models and features is high, we provided a visualization tool that app developers can use to try different classes of these models and visualize, compare the results by three metrics of F1-score, model accuracy, and the run time. 
    
    Further, we study approaches to investigate the energy related reviews in more detail. We compare the results of identifying energy consumption related issues using regular expressions and string matching with two state-of-the-art topic modeling algorithms, online Biterm Topic Modeling (oBTM) \cite{OBTM} and Adaptive Online Latent Dirichlet Allocation (AOLDA) \cite{AOLDA}. The comparison will show the number of distinct issues that can be identified with each approach. Authors have worked in close collaboration with four developers in the identification and evaluation of the string matching and topic modeling results.
    
    The results of this paper and the visualization tool can help the developers or researchers in model and feature selection when building automatic tools for investigating energy related issues reported by users. 
    
    The rest of this paper is organized as follows: In section II, we explain the methodology and study design, followed by reporting the experiments results in Section III. We discuss the results and limitations of the study in Sections IV and V. The related works are summarized in Section VI, and we conclude in Section VII.

\section{Methodology and Study Design}
In this section, we discuss our research questions and explain the methodology, study design in details.

\subsection{Research Questions and Preliminary Result}

The goal of this work is to identify the optimal, top-performing supervised approach to help developers extract energy efficiency related user feedback automatically. Furthermore, we want to investigate if different topic modeling algorithms can help the developers identify  unforeseen issues behind energy efficiency. We, therefore, work our way to answer following research questions:

\begin{enumerate}
    \item[\textbf{RQ1.}] Can we replace manual search for extracting energy related reviews with decent accuracy? 
    
    \textit{Ans: Some Deep Learning approaches equipped with text-enriching techniques performed quite well.}
    
    \item[\textbf{RQ2.}] For smaller training dataset, how does traditional machine learning models perform with respect to comparatively modern NN based models? 
    
    \textit{Ans: With proper feature engineering some traditional ML models beat many of the Deep Learning approaches.}
    
    \item[\textbf{RQ3.}] Is there any opportunity cost incurred by the better performing approaches? 
    
    \textit{Ans: Significantly higher run time is recorded for the better performing approaches.}
    
    \item[\textbf{RQ4.}] To what extent can modern topic modeling algorithms help developers to discover recently emerged issues responsible for energy inefficiency of the application? 
    
    \textit{Ans: Modern topic modeling algorithms can quickly identify major energy related issues (if there is any); but developers can not rely on them to discover \underline{all} the issues.}
\end{enumerate}


\subsection{Methodology}
\begin{figure}[h]
  \centering
  \includegraphics[width=\linewidth]{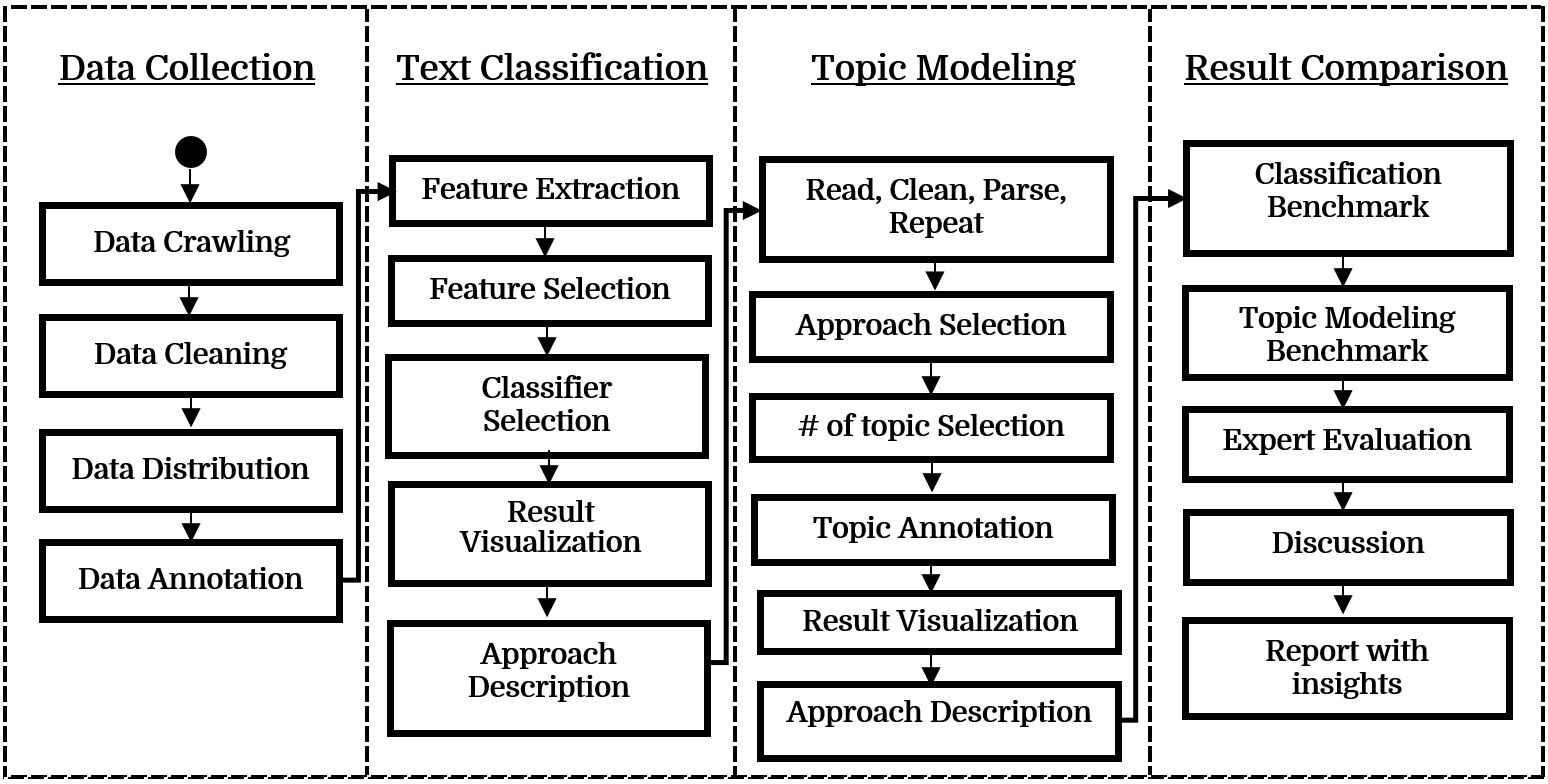}
  \caption{Overview of the study design}
  \label{fig:studyDesign}
\end{figure}

Figure \ref{fig:studyDesign} demonstrates the main steps of our methodology. We have divided the whole procedure of our experiment into smaller steps and assembled them into 4 different groups of \textit{Data Collection}, \textit{text classification}, \textit{Topic Modeling}, \textit{Result Comparison} (represented by 4 different columns). 4 participating developers have carefully followed through each of these steps to report back empirical data, insights presented in the paper.

The first part of \textit{Data Collection} consists of the steps concerning how we collected the reviews by crawling Google play-store and distributed them among the participating developers who later on annotated the reviews (described in \ref{sec:sd}). 
In the \textit{text classification} part, we included the steps for evaluating performances of different machine learning approaches (with engineered features) and different neural network architectures (with pretrained word embedding) (described in \ref{sec:tc}). Here, steps of feature engineering, selection, use of traditional machine learning models for classifier design are discussed in details.
For the \textit{text classification} part, steps involved in energy related issues identification such as: text-processing, algorithm selection, annotating issues to generated topics are included. (\ref{sec:tm}). Here, we evaluated performance of different topic modeling algorithms to generate topics that could illuminate emerging issues discussed in the reviews.
In the final part of \textit{Result Comparison}, steps for reporting on the results of the classification approaches and topic modeling experiments are included (discussed in \ref{sec:rs}).

\subsection{Study Design}
\label{sec:sd}
To ensure involving the experts’ opinions in the study, and reduce the bias in the results that might come from the authors’ previous experience with app reviews automatic text classification, we performed our study with the participation of four Android app developers. Initial guidelines and gold standards (for labeling the energy related reviews) are provided to the developers. After collecting and cleaning the data, the reviews are distributed among the developers according to Table \ref{table:dist}. Each of the developers performed all steps shown in Figure \ref{fig:studyDesign} using different set of reviews from different apps. After the completion of the tasks from each column developers reported back their results in tabular form and based on that data, we developed and provided unique visualization tool to the developers for easy exploration of the result set.
\hfill\\
\textbf{Data Distribution and Annotation:}
\hfill\\
We have collected 100,000 most recent English reviews from 12 different categories. 
20 top-free and 20 top-paid applications have been randomly selected to scrape the reviews.
We have used an \textit{open source Node.js module} \footnote{https://github.com/facundoolano/google-play-scraper} to scrape application data from the Google Play store. It retrieves the full detail of specified application, and maximum number of reviews for that application allowed by Google.

From the data collected, we randomly sampled 40,000 reviews. We have distributed the reviews among 4 participating developers in such a way so that no two developers have same reviews from the same app and the reviews in each developer’s dataset is unique (not shared with others; although developers may discuss the reviews while annotating). Details of reviews distribution among the developers are specified in Table \ref{table:dist}. The developers have discreetly chosen 1,200 reviews each, where 600 reviews were energy-related and the rest were not energy consumption related.

The reason behind choosing such small review- set for each developer is that, usually manual- analysis for investigating energy related reviews are done most frequently by the developer for single application and the accumulated review- size is never very large.

\begin{table}[h!]
\vspace{-2mm}
\caption{Reviews distribution among 4 of the developers}
\label{table:dist}
\begin{center}
\begin{tabular}{p{1.8cm}||p{1.7cm}|p{1.6cm}|p{1.8cm}} 
\hline
\textbf{Developer Aliases} &\textbf{Number of Apps} &\textbf{Number of Reviews} &\textbf{Number of Annotations}\\
  \hline
  \multirow{2}{6em}{Developer-1} & 30 (top free) & 6,540 &  830\\ 
  & 40 (top paid) & 3,504 & 370 \\
  \hline
  \multirow{2}{6em}{Developer-2} & 60 (top free) & 6,854 &  710\\ 
  & 40 (top paid) & 2,512 & 490 \\
  \hline
  \multirow{2}{6em}{Developer-3} & 70 (top free) & 7,584 &  605\\ 
  & 40 (top paid) & 3,028 & 595 \\
  \hline
  \multirow{2}{6em}{Developer-4} & 40 (top free) & 6,012 &  635\\ 
  & 80 (top paid) & 3,966 & 565 \\
  \hline
  \textbf{Total} & \textbf{400} & \textbf{40,000} & \textbf{4,800}\\
  \hline
\end{tabular}
\end{center}
\vspace{-1.5mm}
\end{table}

The developers were instructed to label each review to exactly one class. In case of any confusion during labelling, a review can be openly discussed among the developers and annotated with the agreement of at least two, three in case of a disagreement. The process of annotations were completely manual and all the labelling were thoroughly double checked by 5 different undergraduate students (who were paid) in the following manner: we divided 4,800 reviews into 48 batches of 100 reviews; from each batch, at random 5 reviews are selected and checked. If 1 erroneous labelling is found then that particular batch returns to the respective developer to start its' journey back from stage 1. Out of 48 batches, we have found 2 errors in a single batch only.

\hfill\\
\textbf{Replication Package}
\hfill\\
To encourage reproducibility, we uploaded all scripts, visualization code and benchmark results in a \textit{gitlab account} \footnote{https://gitlab.com/Mashuk/acm-msr-2020-empirical-study-for-text-classification-and-modeling-for-energy-consumption-related-reviews-in-google-play-store.git} and will provide the annotated dataset upon request.

\section{Experiment}
\subsection{Text Classification}
\label{sec:tc}
For autonomous extraction of energy related reviews, we empirically studied 60 machine learning models with various sets of feature combinations and 30 deep learning models with six architectures using three pretrained word embeddings. 
Each developer splitted their review set into training and validation sets with the ratio of 5:1. For both training and validation dataset, we maintained 1:1 ratio for energy related and non energy related reviews. 
\hfill\\
\subsubsection{\textbf{Traditional Machine Learning}}
\hfill\\
\textbf{A. Text Preprocessing}
\hfill\\
To avoid ambiguity, we have converted all the letters of the reviews into lower cases. We have applied this technique at the very beginning as \textit{lowercasing} all the reviews (textual data) is one of the simplest and most effective form of text pre-processing. Although commonly overlooked, it helps significantly when it comes to consistency of expected output. If our dataset contains mixed-case occurrences of any given word (such as "Battery", "battery", "BATTERY"), which it does, it is most likely to effect the features that we are planning to extract from the text reviews if we do not apply \textit{lowercasing}. It is a great way to deal with sparsity issues when the dataset we are working on is rather small like ours. An example of mapping same words with different cases to the same lowercase form is provided: the following words "POWER", "power", "Power" which appear in the raw review text are all converted into "power" using \textit{lowercasing}. 

We have also masked account names, links, and hashtags using the following manner: \textcolor{black}{whenever an account is addressed within the review using the “@” symbol, we took the adjacent word and replace the account name with “account”. We followed the same procedure for hashtags: any word adjacent to "\#" symbols was replaced by word "hashtag". For the links we have used a regular expression filter to find the links and replace it with the word "link".}

We did not remove meaningless words such as "lol", "fugly" since many of them are really popular among mainstream users, yet can not be found in traditional dictionary. These abbreviations, acronyms, and word combinations in reviews appears to be very important as written communication in instant messaging, text messaging, chat, and other forms of electronic communication such as user feedback seemed to have generated a “new language” and contain great weight when it comes to sentiment. \cite{Varnhagen}

After that, we have removed punctuation and stop words from the reviews. \textcolor{black}{Stop words are a set of most frequently used words in a language. Examples of stop words that were removed from our reviews are "i", "me", "my", "myself", "we", "our" etc. These words contain low information about the text and so by removing them we are enabling our features to focus more on the important words.
Let's consider an example in the context of one of our reviews: “It is frustrating... the app is always running in the background... draining my battery really fast”; we would want our features to focus on the following words "frustrating app running background draining battery". If we let our feature extraction procedure to analyze every word in the sentence, words with low information might through the generated features off of its true goal which is detecting energy related reviews. On top of that, preventing all stop words from being analyzed can help us minimize the number of features in consideration while keeping our models decently sized. Instead of replacing the stop words with a dummy character, we ennuled them entirely.}

In the next step, we used \textit{Lemmatization} which necessarily means mapping common words into their base forms. \textcolor{black}{The objective of \textit{Lemmatization} is to remove inflections and map a word to its root form. There is another approach for removing inflection: \textit{stemming} which also reduces the word into its; root form but the “root” in this case may not be a base word, but just a canonical form of the original word. The only difference between \textit{stemming} and \textit{lemmatization} is the latter one tries to accomplish the task in the proper way. \textit{lemmatization} does not just chop prefix/ suffix off, it actually transforms words to the actual root. That is why we have used only \textit{lemmatization} for out purpose. For instance, the words "good", “better”, "best" in the reviews would map to “good” using a dictionary, namely WordNet for mappings.}

\begin{table}
  \caption{Average Accuracy, F1-score, Run-time for most prominent Machine Learning Model+ Feature combinations}
  \label{tab:freq}
  \begin{tabular}{p{1.2cm}||p{2.8cm}|p{1cm}p{0.8cm}p{.8cm}}
    \hline
    \textbf{Model} & \textbf{Feature} & \textbf{Accuracy} & \textbf{F1-score} & \textbf{Run-time}\\
    \hline
    NB & Count Vector-(i) & 0.83& 0.838 & 4.32s\\
    NB & WordLevel TF-IDF-(ii) & 0.84& 0.845 & 4.21s\\
    NB & N-Gram TF-IDF-(iii) & 0.84& 0.837 & 4.22s\\
    NB & CharLevel TF-IDF-(iv) & 0.81& 0.812 & 4.34s\\
    NB & (i)+(ii)$^\dag$ & 0.84& 0.840 & 4.36s\\
    NB & (i)+(iii)$^\dag$ & 0.84& 0.843 & 4.25s\\
    NB & (i)+(iv)$^\dag$ & 0.84& 0.838 & 4.26s\\
    NB & (ii)+(iii)$^\dag$ & 0.87& 0.870 & 4.22s\\
    NB & (ii)+(iv)$^\dag$ & 0.85& 0.847 & 4.88s\\
    NB & (iii)+(iv)$^\dag$ & 0.85& 0.851 & 4.18s\\
    NB & (i)+(ii)+(iii)$^\dag$ & 0.85& 0.846 & 5.91s\\
    NB & (i)+(ii)+(iv)$^\dag$ & 0.84& 0.840 & 5.83s\\
    NB & (i)+(iii)+(iv)$^\dag$ & 0.85& 0.846 & 5.46s\\
    NB & (ii)+(iii)+(iv)$^\dag$ & 0.85& 0.846 & 5.88s\\
    NB & (i)+(ii)+(iii)+(iv)$^\dag$ & 0.85& 0.848 & 6.24s\\
    \hline
    LC & Count Vector-(i) & 0.85& 0.869 & 3.56s\\
    LC & WordLevel TF-IDF-(ii) & 0.87& 0.876 & 3.6s\\
    LC & N-Gram TF-IDF-(iii) & 0.83& 0.836 & 3.58s\\
    LC & CharLevel TF-IDF-(iv) & 0.84& 0.841 & 3.51s\\
    LC & (i)+(ii)$^\dag$ & 0.87& 0.869 & 3.78s\\
    LC & (i)+(iii)$^\dag$ & 0.87& 0.873 & 3.57s\\
    LC & (i)+(iv)$^\dag$ & 0.87& 0.869 & 3.44s\\
    LC & (ii)+(iii)$^\dag$ & 0.89& 0.885 & 3.52s\\
    LC & (ii)+(iv)$^\dag$ & 0.88& 0.876 & 3.41s\\
    LC & (iii)+(iv)$^\dag$ & 0.88& 0.878 & 3.79s\\
    LC & (i)+(ii)+(iii)$^\dag$ & 0.87& 0.873 & 4.14s\\
    LC & (i)+(ii)+(iv)$^\dag$ & 0.87& 0.869 & 4.86s\\
    LC & (i)+(iii)+(iv)$^\dag$ & 0.88& 0.875 & 4.25s\\
    LC & (ii)+(iii)+(iv)$^\dag$ & 0.88& 0.875 & 4.36s\\
    LC & (i)+(ii)+(iii)+(iv)$^\dag$ & 0.87& 0.874 & 5.89s\\
    \hline
    SVM & Count Vector-(i) & 0.85& 0.848 & 18.12s\\
    SVM & WordLevel TF-IDF-(ii) & 0.88& 0.880 & 18.32s\\
    SVM & N-Gram TF-IDF-(iii) & 0.88& 0.880 & 18.48s\\
    SVM & CharLevel TF-IDF-(iv) & 0.85& 0.851 & 18.32s\\
    SVM & (i)+(ii)$^\dag$ & 0.85& 0.850 & 18.65s\\
    SVM & (i)+(iii)$^\dag$ & 0.85& 0.850 & 18.28s\\
    SVM & (i)+(iv)$^\dag$ & 0.85& 0.850 & 18.31s\\
    SVM & (ii)+(iii)$^\dag$ & 0.89& 0.885 & 18.27s\\
    SVM & (ii)+(iv)$^\dag$ & 0.88& 0.876 & 18.38s\\
    SVM & (iii)+(iv)$^\dag$ & 0.88& 0.880 & 18.43s\\
    SVM & (i)+(ii)+(iii)$^\dag$ & 0.85& 0.852 & 19.54s\\
    SVM & (i)+(ii)+(iv)$^\dag$ & 0.85& 0.851 & 19.18s\\
    SVM & (i)+(iii)+(iv)$^\dag$ & 0.85& 0.852 & 19.15s\\
    SVM & (ii)+(iii)+(iv)$^\dag$ & 0.85& 0.852 & 19.17s\\
    SVM & (i)+(ii)+(iii)+(iv)$^\dag$ & 0.85& 0.853 & 20.5s\\
  \hline
\end{tabular}
\begin{tabbing}
$^\dag$Combinations of Count Vector, WordLevel, N-Gram, and Char\\Level TF-IDF
\end{tabbing}
\end{table}
\hfill\\
\textbf{B. Feature Engineering}
\hfill\\
Text representation is a fundamental process for information processing which includes the tasks of determining the index terms for documents and producing the numeric vectors corresponding to the documents.
Thus, we have extracted flat feature vectors from text reviews. We implemented \textit{Count Vectors, TF-IDF (Word level, N-Gram level, Character level) Vectors} to obtain relevant features from our dataset.

\textit{Count Vectors} counts the number of occurrences of each word appeared in a review. It is a matrix notation of the dataset in which every row represents a review from the dataset, every column represents a word from the dataset, and every cell represents the frequency count of a particular word in a particular review. 

On the other hand, \textit{TF-IDF} score represents the relative importance of a word in the review and the entire review- set. TF-IDF(Term Frequency- Inverse Document Frequency) score is composed by two terms: Term Frequency computes the normalized frequency of each word, and Inverse Document Frequency computes as the logarithm of the number of the reviews in the corpus divided by the number of reviews where the specific term appears.

TF(t) = (Number of times word t appears in a review) / (Total number of words in the document)

IDF(t) = ln(Total number of documents / Number of documents with term t in it)

These TF-IDF Vectors can be generated at different levels of input tokens such as words, characters, and n-grams
\begin{itemize}
    \item \textit{ Word Level TF-IDF:} Matrix representing tf-idf scores of every word in different reviews
    \item \textit{N-gram Level TF-IDF:} N-grams are the combination of N number of words together. This Matrix is representing tf-idf scores of N-grams. \textit{ngram\_range} is set as \textit{(2,3)} for our experiment.
    \item \textit{Character Level TF-IDF:}  Matrix representing tf-idf scores of character level n-grams in the review- set
\end{itemize}

\textcolor{black}{We have adopted different variations of TF-IDF in our review classification based on the study that Zhang, Yoshida, and Tang presented on the performances of TF*IDF, LSI and multi-word in 2009 (performance were examined on the tasks of text classification). They have investigated the traditional indexing methods as TF-IDF, LSI (latent semantic indexing) with multi-word (which contains more contextual semantics than individual words and possesses favorable statistical characteristics). The performances of TF*IDF, LSI and multi-word were examined on the tasks of text classification, much like ours. Their experimental results demonstrated that TF*IDF and multi-word are comparable when they are applied for Text Classification and LSI was the poorest one of them. Even the rescaling factor of LSI had an insignificant influence on its effectiveness on text classification. \cite{tfidf}}

In Table \ref{tab:freq}, \ref{tab:xgb} and \ref{tab:deep}, we have denoted Count Vector, WordLevel TF-IDF, N-Gram TF-IDF, and CharLevel TF-IDF as (i), (ii), (iii), (iv) respectively. We have combined these features using horizontal stacking and denoted their combination with '+' sign in the tables.

\begin{table}
  \caption{Accuracy, F1-score, Runtime for Different Model and Feature combination for Extreme Gradient Boosting}
  \label{tab:xgb}
  \begin{tabular}{p{1cm}||p{3cm}|p{1cm}p{0.8cm}p{.8cm}}
    \hline
    \textbf{Model} & \textbf{Feature} & \textbf{Accuracy} & \textbf{F1-score} & \textbf{Run-time}\\
    \hline
    XGB & Count Vector-(i) & 0.81& 0.811 & 14.85s\\
    XGB & WordLevel TF-IDF-(ii) & 0.80& 0.809 & 14.81s\\
    XGB & N-Gram TF-IDF-(iii) & 0.74& 0.743 & 14.83s\\
    XGB & CharLevel TF-IDF-(iv) & 0.81& 0.813 & 14.75s\\
    XGB & (i)+(ii) & 0.81& 0.806 & 14.71s\\
    XGB & (i)+(iii) & 0.82& 0.818 & 14.79s\\
    XGB & (i)+(iv) & 0.83& 0.825 & 14.73s\\
    XGB & (ii)+(iii) & 0.81& 0.812 & 14.46s\\
    XGB & (ii)+(iv) & 0.82& 0.824 & 14.33s\\
    XGB & (iii)+(iv) & 0.82& 0.823 & 14.39s\\
    XGB & (i)+(ii)+(iii) & 0.81& 0.813 & 15.31s\\
    XGB & (i)+(ii)+(iv) & 0.82& 0.818 & 15.35s\\
    XGB & (i)+(iii)+(iv) & 0.83& 0.826 & 15.58s\\
    XGB & (ii)+(iii)+(iv) & 0.83& 0.826 & 15.59s\\
    XGB & (i)+(ii)+(iii)+(iv) & 0.82& 0.821 & 15.51s\\
  \hline
\end{tabular}
\vspace{-2.5mm}
\end{table}

\hfill\\
\textbf{C. Experiment Configuration}
\hfill\\
Each participating developer were instructed to train 3 traditional machine learning models: Naive Bayes, Linear Classifier (Logistic Regression), Support Vector Machine using their designated set of reviews. For each model, developers have tried different feature combinations (total 15 of them) as described in the previous section. We have used 4000 \textit{max\_iter} and Limited-memory Broyden–Fletcher–Goldfarb–Shanno algorithm (L-BFGS) as the solver for Logistic Regression. For SVM, gamma scaling was set to \textit{true}. Python library Keras \cite{keras} was used for composing, training, and evaluating the mentioned models.

In these tables, we have reported the performance of 30 traditional machine learning models with engineered feature combinations using the mean of the following metrics: accuracy, F1-score, and run-time. Once the developers reported back their respective accuracy, F1-score, and run-time, we took the mean value for each case in the following manner.
If reported accuracy, F1-score and run time for a specific model+feature is \textit{x1, y1, z1 (for developer-1); x2, y2, z2 (for developer-2); x3, y3, z3 (for developer-3); and x4, y4, z4 (for developer-4);}
then mean or average accuracy = (x1+x2+x3+x4)/4, mean F1-score = (y1+y2+y3+y4)/4, and mean Runtime = (z1+z2+z3+z4)/4.  


We have investigated the deviation of each reported values from the average accuracy, F1-score, Run-time and found that for each case maximum deviation calculated were $\pm 0.02$, $\pm 0.021$, and $\pm 1.4s$ respectively. For every model, different combination of TF-IDF vectors prevailed yielding better performance with each of the traditional machine learning models.

\textit{Extreme Gradient Boosting: } In traditional machine learning, ensemble methods are quite popular as they use multiple learning algorithms to obtain better predictive performance than could be obtained from any of the constituent learning algorithms alone \cite{ensemble1, ensemble2, ensemble3}. A machine learning ensemble consists of only a concrete finite set of alternative models, but typically allows for much more flexible structure to exist among those alternatives.

Boosting is a popular ensemble method which involves incrementally building an ensemble by training each new model instance to emphasize the training instances that previous models misclassified. In most cases of text categorization (like ours), boosting has been shown to yield better accuracy than bagging, but it also tends to be more likely to over-fit the training data. \cite{ensembleWiki}
To include ensemble method into our tool we have our developers train an Extreme Gradient Boosting Model (XGB) with all 15 feature combinations and presented the data for XGB in Table \ref{tab:xgb}. 
\hfill\\
\subsubsection{\textbf{Deep Learning}}
\hfill\\
We have tried better feature engineering for getting improved performance out of traditional machine learning models. For deep learning, we will try a text enrichment approaches such as using word embeddings.
\hfill\\
\hfill\\
\textbf{A. Word Embedding}
\hfill\\
We leveraged pre-trained word embedding (an approach for representing words and documents using dense vector representation) to work with our Neural Network based models. Traditional bag-of-word model encoding schemes were the most popular and largely used before word embedding came into light. For a predefined vocabulary of fixed length (derived from a corpus of text such as our review-set), word embedding methods learn a real-valued vector representation where dense vectors, vectors represent words and the projection of the word into a continuous vector space respectively. Within the vector space, the position of a word is determined from text (specifically from the words that surround the word when it is being used). A word's embedding is referred to as the position of a word in the learned vector space\cite{wordEmbed}. Three most popular examples of carefully designed methods of learning word embeddings from text include:
\begin{itemize}
    \item fastText\cite{fasttext} 
    \item GloVe \cite{glove}, and 
    \item Word2Vec\cite{w2v}.
\end{itemize}

We could train word embeddings using the input corpus itself. Two most popular way to train our own word embedding are \textit{Learn it Standalone} and \textit{Learn Jointly}. The first one is a good approach for using the same embedding in multiple models where a model is trained to learn the embedding, saved and later used as a part of another model. To use the embedding on one task, second approach may be used where the embedding is learned as part of a large task-specific model. But learning a word embedding from scratch for our problem would require a large amount of text data (millions of documents containing large number of words) to ensure that useful embeddings are learned.

Word Embeddings could also be downloaded from previously trained models as it is common for researchers to make pre-trained word embeddings available (for free; often under permissive license). For
instance, FastText, Word2Vec and GloVe word embeddings are available for free download. These word embeddings (trained on the same corpus: 1M wiki-news documents) were used on our project instead of training our own embeddings from scratch. 

There are two popular options: \textit{Static} and \textit{Updated} when it comes to using pre-trained embeddings. For the first option the embedding is kept static and can be used as a component of our model and for the second option the pre-trained embedding is used to seed the model, and then the embedding is jointly updated during the training of our model. The second option suits our problem most as we want get the most out of our model and embedding to accomplish the task. \cite{wordEmbed}

Brief discussion about the models (that have been used to train our word embeddings) are given below.

\begin{table}[htbp]
  \caption{Accuracy, F1-score, Runtime for Different Model and Feature combination for  Shallow Neural Network and some Deep Learning Models}
  \label{tab:deep}
  \begin{tabular}{p{1.2cm}p{2.8cm}p{1cm}p{0.8cm}p{.8cm}}
    \hline
    \textbf{Model} & \textbf{Feature} & \textbf{Accuracy} & \textbf{F1-score} & \textbf{Run-time}\\
    \hline
    S-NN & Count Vector-(i) & 0.86& 0.870 & 19.21s\\
    S-NN & WordLevel TF-IDF-(ii) & 0.86& 0.867 & 19.2s\\
    S-NN & N-Gram TF-IDF-(iii) & 0.83& 0.838 & 19.25s\\
    S-NN & CharLevel TF-IDF-(iv) & 0.82& 0.840 & 19.29s\\
    S-NN & (i)+(ii) & 0.87& 0.871 & 19.22s\\
    S-NN & (i)+(iii) & 0.87& 0.874 & 19.21s\\
    S-NN & (i)+(iv) & 0.87& 0.874 & 19.23s\\
    S-NN & (ii)+(iii) & 0.90& 0.889 & 19.29s\\
    S-NN & (ii)+(iv) & 0.88& 0.877 & 19.25s\\
    S-NN & (iii)+(iv) & 0.88& 0.876 & 19.72s\\
    S-NN & (i)+(ii)+(iii) & 0.88& 0.877 & 20.75s\\
    S-NN & (i)+(ii)+(iv)& 0.87& 0 .87 & 20.76s\\
    S-NN & (i)+(iii)+(iv) & 0.88& 0.877 & 20.45s\\
    S-NN & (ii)+(iii)+(iv) & 0.88& 0.882 & 20.52s\\
    S-NN & (i)+(ii)+(iii)+(iv) & 0.89& 0.878 & 21.85s\\
    \hline
    CNN & WE-FastText* & 0.94 & 0.935 & 35.28s\\
    CNN & WE-Glove* & 0.91 & 0.892 & 35.23s\\
    CNN & WE-Word2Vec* & 0.89 & 0.901 & 35.81s\\
    \hline
    RNN\_LSTM & WE-FastText* & 0.89 & 0.887 & 42.38s\\
    RNN\_LSTM & WE-Glove* & 0.85 & 0.854 & 42.49s\\
    RNN\_LSTM & WE-Word2Vec* & 0.81 & 0.811 & 42.16s\\
    \hline
    RNN\_GRU & WE-FastText* & 0.84 & 0.829 & 36.19s\\
    RNN\_GRU & WE-Glove* & 0.86 & 0.857 & 36.28s\\
    RNN\_GRU & WE-Word2Vec* & 0.81 & 0.814 & 36.17s\\
    \hline
    Bidir\_RNN & WE-FastText* & 0.86 & 0.864 & 41.54s\\
    Bidir\_RNN & WE-Glove* & 0.87 & 0.864 & 41.14s\\
    Bidir\_RNN & WE-Word2Vec* & 0.85 & 0.858 & 41.19s\\
    \hline
    RCNN & WE-FastText* & 0.90 & 0.908 & 37.85s\\
    RCNN & WE-Glove* & 0.90 & 0.891 & 37.66s\\
    RCNN & WE-Word2Vec* & 0.92 & 0.935 & 37.25s\\
    
  \hline
\end{tabular}
\begin{tabbing}
* Word Embeddings trained with fastText, Glove, and Word2Vec \\are referred to as WE-FastText, WE-Glove, and WE-Word2Vec \\respectively 
\end{tabbing}
\vspace{-2mm}
\end{table}

\textit{Word2Vec\cite{w2v}} takes a large corpus as its input and produces a vector space (typically of several hundred dimensions) with each unique word in the corpus being assigned a corresponding vector in the space. Word vectors are positioned in the vector space such that words that share common contexts in the corpus are located close to one another in the space. \textcolor{black}{In 2013, after Mikolov, Chen, Corrrado, and Dean published \cite{w2v1}, the avalanche of word embeddings began. The proposed approach (Word2Vec) uses small neural networks to calculate word embeddings based on the context of the words. The authors of the paper published two approaches to implement this: Continuous Bag of Words (CBOW) and skip-gram. In the first approach, for the given context, the network tries to predict which word is most likely to occur/appear. The network predicts a similar probability for words that are equally likely to appear (which can be interpreted as having a shared dimension).
For the second approach, the network works the other way around, that is, uses the target word to predict its context. The results of Word2Vec were unprecedented but also hard to explain from a theoretical point of view.}

\begin{figure*}[t]
  \includegraphics[width=\linewidth]{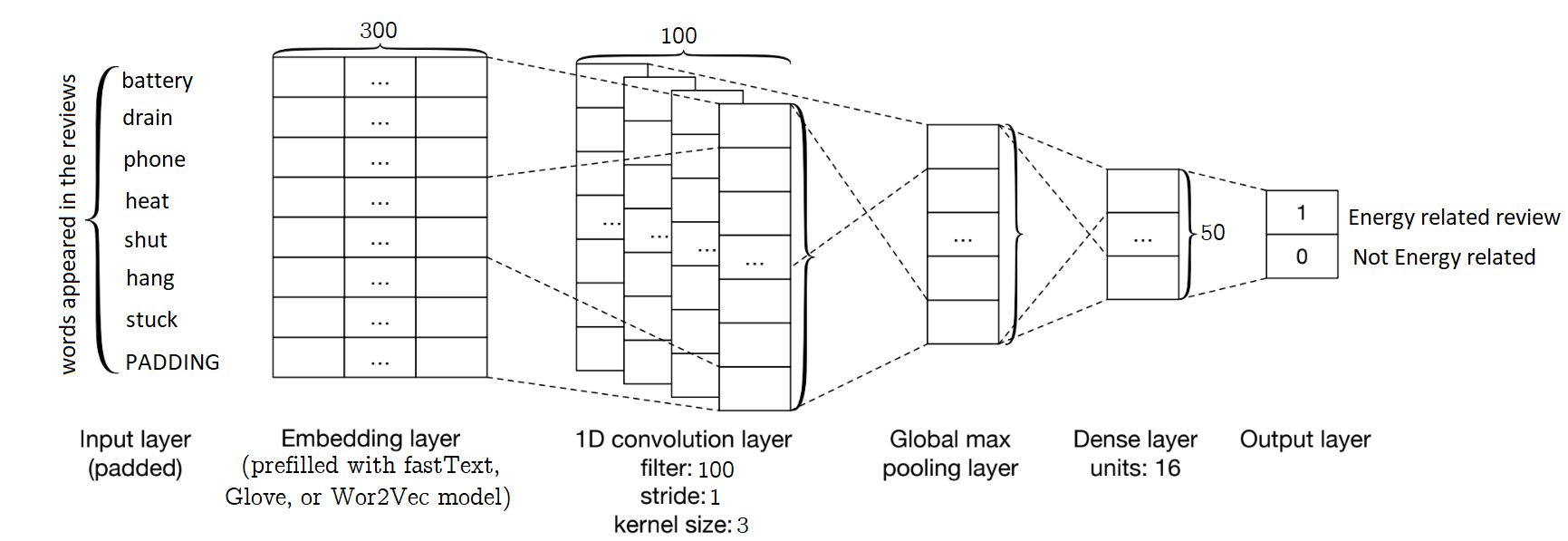}
  \caption{One of the Neural Network Architecture (CNN) used for text classification (yielded best performance)}
  \label{fig:cnn}
\end{figure*}

\textit{GloVe \cite{glove}} is an unsupervised learning algorithm which obtains vector representations for words by mapping words into a meaningful space where the distance between words is related to semantic similarity. \textcolor{black}{One year after the publication of Word2Vec\cite{w2v1}, researchers of Stanford published GloVe\cite{glv1}. Glove is deemed as a slight variation of Word2Vec which attempts to benefit from a less obvious aspect of Word2Vec.
Word2Vec learns embeddings by relating target words to their context, ignoring co-occurrences of some context words as if the frequent co-occurrence of words only helps it creating more training examples, but carries no additional information.
On the other hand, GloVe accentuates on the frequency of co-occurrences and takes it as vital information rather than wasting it as mere additional training examples. So, GloVe focuses on building word embeddings in such a way that the probability of words’ co-occurrence in the corpus relates directly to a combination of corresponding word vectors.
In brief, its embeddings can be interpreted as a summary of the training corpus with lower rank in dimensions that associates co-occurrences.}

\textit{fastText\cite{fasttext}} is formally a library for learning of word embeddings and text classification created by Facebook's AI Research (FAIR) lab\cite{ft1, ft2, ft3, ft4}. The model allows to create an unsupervised learning or supervised learning algorithm for obtaining vector representations for words. Facebook makes available pretrained models for 294 languages.\cite{ft5} \textit{fastText} uses a neural network for word embedding.is an unsupervised approach to learn high dimensional vector representations for words from a large training corpus where the vectors of words that occur in a similar context are close in this space.\cite{ftWiki}
\textcolor{black}{\textit{fastText}'s  kept its' base idea pretty similar to Word2Vec where instead of using words to build word embeddings, it goes one level deeper and takes characters, part of words as the building block for the embedding so that a word becomes its own context.
Word embeddings generated by FastText and Word2Vec are almost similar although they are not calculated directly. Instead, they are generated as a combination of lower-level embeddings.
Two main advantages to this approach are \textit{generalization} and \textit{necessity of less training data}. Generalization comes to play when there are new words which have the same characters as known ones. Additionally, we need less training data as much more information can be extracted from each piece of document.}

For our study purpose, we downloaded three word embeddings; all of three embeddings were generated through training same corpus of 1 million wiki-news with three aforementioned models: fastText, Glove. Word2Vec. To make each of these word embeddings applicable with Neural Network based models, we implemented following procedure: first we loaded a pre-trained word-embedding vectors in order create a tokenizer, then converted text to sequence of tokens, padded them to ensure equal length vectors, and at last created token-embedding mapping in the form of \textit{embedding matrix}.

\hfill\\
\textbf{B. Neural Network Architectures}
\hfill\\
Traditional machine learning models depends on a data representation relying upon hand-crafted features, chosen by users or domain experts, based on its' usefulness for the classification problem at hand. On the other hand, neural networks (used in deep learning approaches) learn high-level feature representations automatically by using raw textual data as input.
Researchers have previously applied Convolutional Neural Network (CNN) successfully to natural language processing problems such as text classification\cite{cnn1}. Usually, deep learning approaches require a large amount of training data to outperform traditional machine learning approaches\cite{cnn2}. As our dataset is rather small for conventional deep learning approaches, we are trying to leverage pretrained word embeddings as discussed in the previous section.

Figure \ref{fig:cnn} shows a simple architecture of neural network which was used by Haring et al. \cite{cnnarc} to identify user comments on online news sites that address three different classes: media house, journalist, or forum moderator. We are going to use this architecture and its' different variation for empirical study purpose. 

The input layer for this architecture requires its' textual inputs to be of fixed size; so, we searched for the longest review in our dataset and assigned its' length as the fixed size of input for the specified layer. To make the other reviews reach the required length for the input layer, we have padded them. 
Right after the input layer,  our network consists of an embedding layer which contains the pre-trained word embeddings: we have assigned the embedding matrix (discussed broadly in the previous section) as the weights of this embedding layer and set the \textit{trainable} parameter for this layer to be \textit{false} so that the weights of the embedding layer stays frozen during training. After this embedding layer, our network consists of a 1D convolution layer, a 1D global max pooling layer, a dense layer with a concluding output layer. We have used \textit{ReLU} (Rectified Linear Unit) as the activation function for convolutional layer and \textit{sigmoid} function for the dense output layer.

In Table \ref{tab:deep}, output of this architecture is denoted by CNN. We have swapped the convolution and max pooling layer of described architecture with layers of RNN- LSTM (Recurrent Neural Network– Long Short Term Memory), RNN- GRU (RNN- Gated Recurrent Units), Bidirectional RNN, and RCNN (Recurrent CNN) to try out different variants of this architecture. Participating developers trained the models with a batch size of 32 and 10 epochs. The developers experimented with three mentioned word embeddings and reported the results. For comparison purpose, we also reported the performance for S- NN (Shallow Neural Network) trained with different feature combinations engineered for traditional machine learning models.

The detailed evaluation is presented in Table \ref{tab:deep}. Similar as before, the average accuracy, F1-score and Run-time for 4 of our participating developer is reported for each case.

\begin{figure}[h]
  \centering
  \includegraphics[scale=0.3]{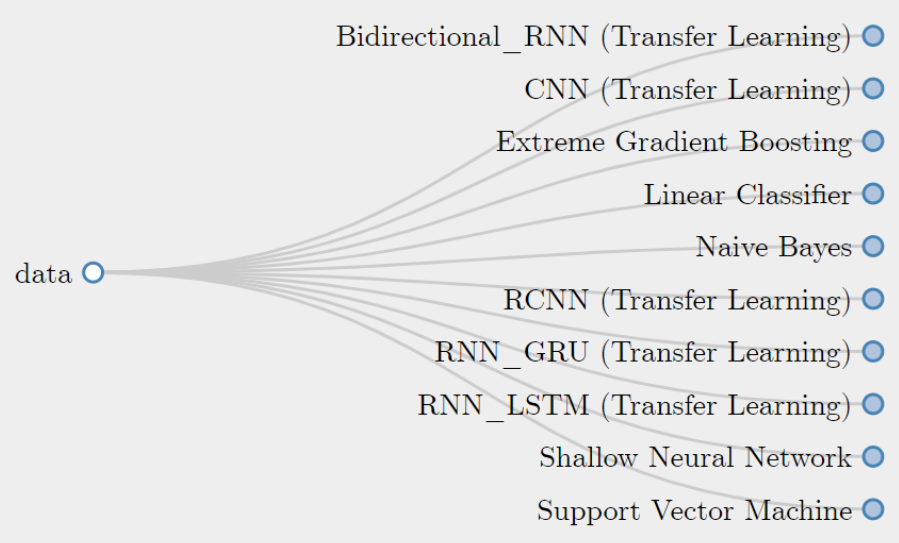}
  \caption{Data Visualization for text classification result set}
  \label{fig:data_viz}
\vspace{-4mm}
\end{figure}

\hfill\\
\textbf{C. Data Visualization}
\hfill\\
We developed an interactive dynamic tree where each node in the first, second and third layer represent different models, feature, and result-data (accuracy, F1-score, run time) respectively. Each node can be dragged, zoomed and panned. This part of the tool was developed so that developers may traverse through the resultant data set with ease. In figure \ref{fig:data_viz}, we have only shown up to the first level of model selection where each node can expand into different number of second level nodes to accommodate user to select different features and generate the results. 

\begin{figure*}[t]
  \includegraphics[width=\linewidth]{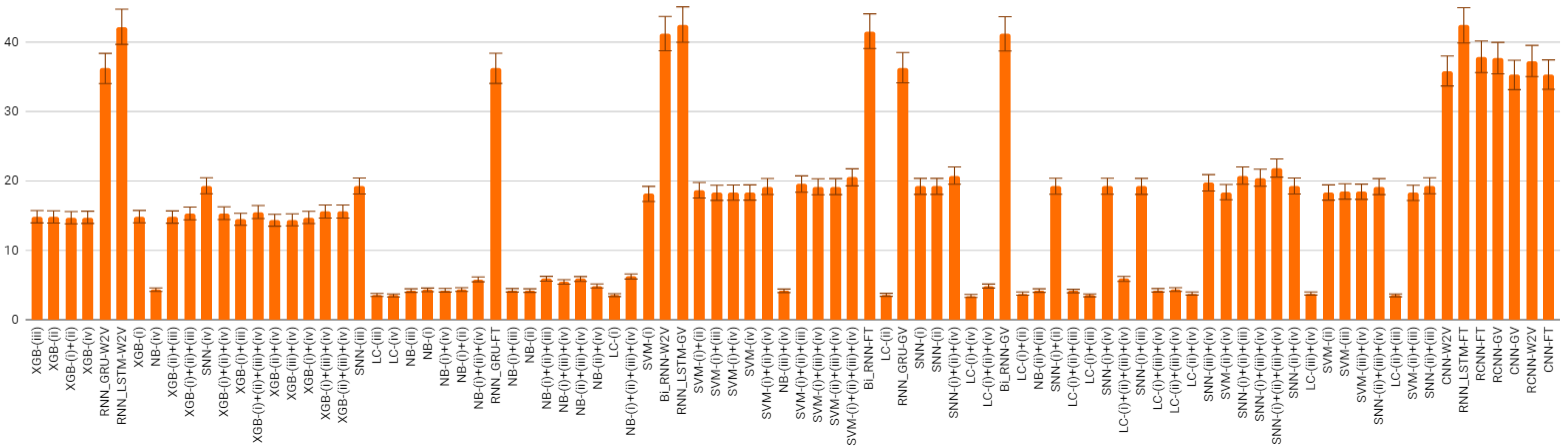}
  \caption{Average Runtime of developers' trials for efficient text-classification}
  \label{fig:run_time}
\end{figure*}

\hfill\\
\textbf{D. Runtime- Performance Trade off}
\hfill\\
Figure \ref{fig:run_time} shows the average run time for all 90 cases described in the previous traditional Machine Learning and Deep Learning sections. In the x- axis, all the model feature combinations are presented in the ascending order (left-to-right) of their F1-score performance. Here, we observed that although Neural Network based models are outperforming traditional machine learning models accuracy-wise, these NN based models require significantly greater run-time than those of traditional ML models. Run time is almost perceived as an opportunity cost for accuracy or F1-score. When it comes to industrial implementation, developers may choose to use a model+feature combination that might yield sub-optimal accuracy but run faster than other choices.

\begin{table}
  \caption{Average Accuracy, F1-score, Run-time for 32 most prominent Machine Learning Model+ Feature combinations}
  \label{table:comparison}
  \begin{tabular}{p{1.2cm}p{2.8cm}||p{1cm}p{.8cm}p{.8cm}}
    \hline
    \textbf{Model} & \textbf{Feature} & \textbf{Accuracy} & \textbf{F1-score} & \textbf{Run-time}\\
    \hline
    NB & (ii)+(iii)$^\dag$ & 0.86& 0.861 & 5.59s\\
    NB & (ii)+(iv)$^\dag$ & 0.85& 0.848 & 5.41s\\
    NB & (i)+(iii)+(iv)$^\dag$ & 0.84& 0.844 & 5.27s\\
    NB & (i)+(ii)+(iii)+(iv)$^\dag$ & 0.83& 0.838 & 5.64s\\
    \hline
    LC & (ii)+(iii)$^\dag$ & 0.89& 0.891 & 4.85s\\
    LC & (ii)+(iv)$^\dag$ & 0.88& 0.879 & 4.91s\\
    LC & (i)+(iii)+(iv)$^\dag$ & 0.87& 0.873 & 5.36s\\
    LC & (ii)+(iii)+(iv)$^\dag$ & 0.87& 0.872 & 5.12s\\
    \hline
    SVM & (ii)+(iii)$^\dag$ & 0.89& 0.887 & 19.34s\\
    SVM & WordLevel TF-IDF-(ii) & 0.88& 0.884 & 18.32s\\
    SVM & N-Gram TF-IDF-(iii) & 0.88& 0.882 & 18.48s\\
    SVM & (iii)+(iv)$^\dag$ & 0.88& 0.883 & 19.85s\\
    \hline
    \hline
    XGB & (i)+(iv)$^\dag$ & 0.82& 0.824 & 14.49s\\
    XGB & (i)+(iii)+(iv)$^\dag$ & 0.81& 0.818 & 14.52s\\
    XGB & (ii)+(iii)+(iv)$^\dag$ & 0.80& 0.812 & 14.54s\\
    \hline
    \hline
    S-NN & Count Vector-(i) & 0.86& 0.870 & 18.21s\\
    S-NN & WordLevel TF-IDF-(ii) & 0.86& 0.867 & 18.2s\\
    S-NN & N-Gram TF-IDF-(iii) & 0.83& 0.838 & 18.25s\\
    S-NN & CharLevel TF-IDF-(iv) & 0.82& 0.840 & 18.29s\\
    S-NN & (ii)+(iii)$^\dag$ & 0.90& 0.889 & 18.19s\\
    \hline
    \hline
    CNN & WE-FastText* & 0.93 & 0.939 & 34.17s\\
    CNN & WE-Glove* & 0.92 & 0.895 & 36.12s\\
    CNN & WE-Word2Vec* & 0.90 & 0.902 & 33.1s\\
    \hline
    RNN\_LSTM & WE-FastText* & 0.89 & 0.884 & 41.12s\\
    RNN\_LSTM & WE-Glove* & 0.86 & 0.853 & 43.15s\\
    RNN\_LSTM & WE-Word2Vec* & 0.80 & 0.821 & 42.12s\\
    \hline
    RNN\_GRU & WE-FastText* & 0.83 & 0.829 & 34.25s\\
    RNN\_GRU & WE-Glove* & 0.85 & 0.847 & 34.25s\\
    RNN\_GRU & WE-Word2Vec* & 0.82 & 0.834 & 35.27s\\
    \hline
    Bidir\_RNN & WE-FastText* & 0.87 & 0.854 & 42.48s\\
    Bidir\_RNN & WE-Glove* & 0.87 & 0.867 & 41.25s\\
    Bidir\_RNN & WE-Word2Vec* & 0.84 & 0.850 & 40.15s\\
    \hline
    RCNN & WE-FastText* & 0.89 & 0.901 & 37.46s\\
    RCNN & WE-Glove* & 0.90 & 0.896 & 35.48s\\
    RCNN & WE-Word2Vec* & 0.91 & 0.925 & 37.25s\\
    \hline
\end{tabular}
\begin{tabbing}
$^\dag$Combinations of Count Vector, WordLevel, N-Gram, and Char\\Level TF-IDF; 
* Word Embeddings trained with fastText, Glove, \\and Word2Vec are referred to as WE-FastText, WE-Glove, and \\WE-Word2Vec respectively 
\end{tabbing}
\vspace{-6mm}
\end{table}

\hfill\\
\textbf{E. Validity of the approach}
\hfill\\
To ensure that the designed approaches and the result is valid and analogous to other binary classification cases, authors performed a similar experiment to compare the outcomes without developers involvement. We selected 600 security related application reviews and 600 non- security related reviews and applied the same approach developers used for automatic extraction of the energy related reviews. All the 90 instances (60 traditional machine learning approach, 30 Neural Network based approach) were observed but no significant difference in the accuracy, F1-score and run time. Maximum difference reported is $\pm .03$ for both accuracy and F1-scores. For run-time maximum difference recorded was $\pm 3.9s$. For brevity, we have reported 32 most prominent cases out of 90 in Table \ref{table:comparison}.

\subsection{Topic Modeling}
\label{sec:tm}
After the autonomous separation of the energy-related reviews, we need to investigate the reviews to discover issues that are causing energy inefficiency. Previously, developers have manually searched for suspected issues using manual inspection; i.e. "string matching". We want to see if we can find these issues using topic modeling. We have provided two different approaches for topic modeling, one is \textit{Adaptive Online Latent Dirichlet Allocation} 
and the other one is \textit{Online Biterm Topic Modeling}
We compare them with previously used \textit{manual inspection} and present the result.

\subsubsection{\textbf{Different Approaches for New Issues Identification}} 
\hfill\\
Three techniques used for identification of emerging issues from already separated energy related reviews are briefly discussed here. 

\textbf{A. String Matching}

The most used technique to categorize user reviews is to check whether it contains a certain key- word. Developers initially predict what issues might emerge and define a list of keywords to search the reviews using "LIKE" in SQL query. The query in turn return the reviews containing the defined key- words. We compiled the keywords from the literature \cite{rel1} and used for the string matching classifier. Some of the keywords are as following: \textit{battery, energy, power, charge, affect, consume, hog, deplete, discharge, drain, kill, devour, hot, heat, slow, consume, leak} and 24 more. This technique is widely used but it cannot identify any energy related issues unforeseen by developers.

\textbf{B. Adaptive Online Latent Dirichlet Allocation} \cite{AOLDA}

This is an online Variational Bayes algorithm developed for LDA that can be trained in small batches. While implementing Online LDA, we only need to hold a very small subset of the dataset in memory at a given time.\cite{AOLDA2}. According to \cite{AOLDA}, Adaptive Online LDA finds topic models better and faster than those found with batch VB. The batch VB algorithm assumes that user (in our case, developers) would have the entire training set at the start of training, rendering the approach ineffectual for our experiment as app developers would want to train the model incrementally. AOLDA supports incremental training so that developers can train on a chunk of reviews, then resume training if they receive more reviews from the user (practical for our purpose) without have to retrain on the original chunk of reviews. As it also works good with short text data, we have integrated it in our experiment. 

\textbf{C. Online Biterm Topic Modeling Algorithm} 

OBTM is inspired by the online LDA algorithm proposed in \cite{AOLDA2}, which assumes documents are divided by time slices; and within these time slices, the documents are exchangeable. OBTM tries to fit a BTM model over the data in a time slice and use the counts in current time slice to adjust the hyper parameters for the next time slice. Although we could not use it with the streaming data as conveniently as AOLDA, we have integrated it into our experiment as according to \cite{OBTM}, OBTM out performs LDA, iLDA, and iBTM for topic modeling from a set of short text documents.

\subsubsection{\textbf{Comparison in outcomes}}\hfill\\
For Topic modeling part, each developer worked on their unique dataset but only for energy related reviews (each developer originally had 600 exclusive energy related reviews). 
At first, each of them defines some keywords based on the prediction about probable issues and uses them for string matching (process defined in \textit{String Matching} section); number of identified issues from this process is reported in the 3rd column of Table \ref{tab:tm}.  
In the second phase, we manually investigate the true number of issues appeared in reviews and for each developer's review set, the number of \textit{true reviews} discovered is specified in column 2. Let's assume, k number of issues were discovered for developer-A's review- set. 
Then for the third phase, developer- A deploys OBTM, AOLDA with an instruction to generate k number of topics. 'k' is chosen as threshold since we wanted to find out how many (out of the total number of actual issues=k) issues we can identify from the k number of topics generated by OBTM and AOLDA. After that developer- A will define/label the generated topics and try to match them with \textit{true topics} identified in second phase. Recorded number of matched issues are reported in 4th and 5th column of Table \ref{tab:tm}.

\begin{table}[h!]
\caption{Reviews discovered by different algorithms}
\label{tab:tm}
\begin{center}
\begin{tabular}{p{1.8cm}p{1.7cm}p{1.5cm}p{.8cm}p{.8cm}} 
\hline
\textbf{Developers' aliases} &\textbf{True issues (k)} &\textbf{String Matching} &\textbf{aoLDA} &\textbf{oBTM}\\
  \hline
  Developer-1 & 4 & 1 & 3;(k=4) & 3;(k=4)\\
  Developer-2 & 3 & 1 &  2;(k=3) & 3;(k=3)\\
  Developer-3 & 5 & 2 &  3;(k=5) & 4;(k=5)\\
  Developer-4 & 2 & 1 &  1;(k=2) & 2;(k=2)\\
  \hline
\end{tabular}
\end{center}
\vspace{-6mm}
\end{table}

Here, Developer-1 had identified the following 4 issues: (i) User Interface, (ii) Extraneous work, (iii) Defective Task Allocation, (iv) App Dependency. String matching could discover only (ii); aoLDA discovered (i), (iii), (iv); and oBTM also discovered (i), (iii), (iv).

Developer-2 had identified the following 3 issues: (i) Power Save mode (ii) Internet Connection, (iii) Phone Heat up. String matching could discover only (iii); aoLDA discovered (i), (iii); and oBTM discovered (i), (ii).

Developer-3 had identified the following 5 issues: (i) Slow Animation, (ii) High resolution Videos, (iii) Defective Task Allocation, (iv) No power awareness, (v) Cache mismanagement. String matching could discover only (i), (ii); aoLDA discovered (i), (ii), (iv); and oBTM discovered (i), (ii), (iii), (iv).

Developer-4 had identified the following 2 issues: (i) Internet Connection , (ii) App dependency. String matching could discover only (i); aoLDA discovered (i); oBTM discovered all of them to some extent.

\section{Result}
\label{sec:rs}
We have observed that out of 90 instances, top performing seven instances came from four distinct neural network based architectures equipped with moderately large pretrained word embeddings. Here, CNN based architecture supported by word-embedding previously trained with fastText performed the best with highest F1-score (0.935). But not all NN architectures could outperform traditional machine learning (ML) algorithms. We have noticed that with proper feature engineering, traditional ML models can also go on par with neural network based architectures (e.g. SVM with word-level+N-gram TF-IDF as feature).
Another point to take into account is that the significantly larger run time for NN based architectures which can be seen as opportunity cost for the yielded performance. When developers' frequent implementation is taken into account, NN based architectures can be quite detrimental while some traditional machine learning models such as SVM (or any other run time friendly model feature combination) emerges as more expedient choice.
For topic modeling, we observed that topic modeling algorithms have performed quite well than conventional \textit{string matching} method and quite helpful to identify large number of unforeseen issues responsible for energy inefficiency. Developers' opinion summarizes like following: "Both OBTM and AOLDA performs far better than traditional approach. OBTM generates less number of trivial/ non- relevant keywords than AOLDA. While AOLDA encompasses a lot of information in limited number of topics (making it difficult to define the topics using generated keywords), OBTM works better to isolate and expose crucial issues in same number of topics."

\section{Study Limitation}
For text classification, we excluded result set for less prominent models like \textit{Decision Trees}, \textit{Random Forest} as they yielded less promising results. Different variants of NN based architectures could be introduced by adding layers of \textit{Hierarchical Attention Networks, Bidirectional Recurrent Convolutional Neural Networks}. We excluded NN architectures' performance after implementing hyper-parameter tuning as it does not contribute to significant improvement. We could generate various iteration of the results by letting developers use review sets of different sizes.
Topic model algorithms were not implemented to generate varying number of topics to see how they scale the distribution of issues throughout the topics. In the topic modeling part, we have solely focused on qualitative performance and didnot take run-time into account.

\section{Related Work}
Multiple studies investigated the energy consumption of mobile apps/ software among the developers, identified the best practices, determined the energy patterns, or mined the energy related commits. These studies show the importance of energy consumption of mobile apps both from the developers’ and the users’ perspectives.\cite{rel1, Oliner13, noei2019too}. Data mining and analysis for \textit{tweets}\cite{rel2, rel3}, \textit{Amazon Product Review}\cite{rel5} and \textit{Product reviews and description}\cite{rel4} has focused on requirement engineering too. Stanik et al. \cite{stanik} worked on classifying multilingual user feedback (with single CNN architecture and single word embedding) and before him Maalej et al.\cite{maalej} tried to automatically classify app reviews but did not take Neural Network based architecture into account. Furthermore, Automatically mining product opinions from the web and in generating opinion-based summaries of user reviews were attempted in the following work\cite{tm1, tm2, tm3, tm4, tm5, tm6, tm7, tm8}, but all of them worked with sub-optimal topic modeling algorithms.

\section{Conclusion}
We empirically studied 60 machine learning models-features combinations along with 30 neural network based models equipped with 3 word embeddings. We reported the comparison using accuracy, F1-score and run time as metrics where our focus was to automatically extract the energy related reviews. Our study also exposed an opportunity cost (run time) for achieving better performance using mentioned approaches.
We further compare various topic modeling algorithms, OBTM and AOLDA to automatically investigate the emerging issues responsible for energy inefficiency of the apps. We compared the previously used \textit{string matching} with results obtained from applied techniques and presented the result of a qualitative study performed in collaboration with developers and students to determine their preferences.

\bibliographystyle{IEEEtran}
\bibliography{hadi-terminal}

\end{document}